\documentclass[conference]{IEEEtran}
\IEEEoverridecommandlockouts
\usepackage{tabu}
\usepackage{xcolor}
\usepackage{cite}
\usepackage{multirow}
\usepackage{amsmath,amssymb,amsfonts}
\usepackage{algorithmic}
\usepackage{graphicx}
\usepackage{textcomp}
\graphicspath{{./Figures/}}
\DeclareGraphicsExtensions{.pdf,.jpg,.png,.gif}
\def\BibTeX{{\rm B\kern-.05em{\sc i\kern-.025em b}\kern-.08em
    T\kern-.1667em\lower.7ex\hbox{E}\kern-.125emX}}
    
\DeclareMathOperator{\AP}{AP}
\DeclareMathOperator{\OA}{OA}
\begin{document}

\title{Recent Developments from Attribute Profiles \\ for Remote Sensing Image Classification
}

%\author{%
%	\IEEEauthorblockN{Minh-Tan Pham\IEEEauthorrefmark{2}, S\'ebastien Lef\`evre\IEEEauthorrefmark{2}, 
%	Lorenzo Bruzzone\IEEEauthorrefmark{3},
%	Erchan Aptoula\IEEEauthorrefmark{4},
%	 Name Surname\IEEEauthorrefmark{4}%
%	\thanks{This work is supported by the R\'egion Bretagne grant.}}
%	\IEEEauthorblockA{\IEEEauthorrefmark{2}IRISA - Universit\'e Bretagne Sud, UMR 6074, F-56000 Vannes, France}	
%	\IEEEauthorblockA{\IEEEauthorrefmark{3}Dept. of Information Engineering and Computer Science - University of Trento, I-3812 Trento, Italy}
%	\IEEEauthorblockA{\IEEEauthorrefmark{4}Institute of Information Technologies - Gebze Technical University, 41400 Kocaeli, Turkey}
%\texttt{\{minh-tan.pham,sebastien.lefevre\}@irisa.fr}
%}

\author{
\IEEEauthorblockN{Minh-Tan Pham}
\IEEEauthorblockA{\textit{IRISA - Universit\'e Bretagne Sud} \\
\textit{UMR 6074, F-56000, Vannes, France}\\
\texttt{minh-tan.pham@irisa.fr}}
\\
\IEEEauthorblockN{Erchan Aptoula}
\IEEEauthorblockA{\textit{Institute of Information Technologies} \\
\textit{Gebze Technical University, 41400 Kocaeli, Turkey}\\
\texttt{eaptoula@gtu.edu.tr}}
\and
\IEEEauthorblockN{S\'ebastien Lef\`evre}
\IEEEauthorblockA{\textit{IRISA - Universit\'e Bretagne Sud} \\
\textit{UMR 6074, F-56000, Vannes, France}\\
\texttt{sebastien.lefevre@irisa.fr}}
\\
\IEEEauthorblockN{Lorenzo Bruzzone}
\IEEEauthorblockA{\textit{Dept. of Inf. Eng. Computer Science} \\
\textit{University of Trento, I-3812 Trento, Italy}\\
\texttt{lorenzo.bruzzone@unitn.it}}
%\and
%\IEEEauthorblockN{Name Surname}
%\IEEEauthorblockA{\textit{Institute of Information Technologies} \\
%\textit{Gebze Technical University, 41400 Kocaeli, Turkey}\\
%\texttt{eaptoula@gtu.edu.tr}}
}

\maketitle
%\textcolor{red}{\bf Note that S\'ebastien and I propose this planning which can be applicable for both short and long versions. In the short version, Section III may focus on only some points in details, others come with brief introduction and related references.}

\begin{abstract}
Morphological attribute profiles (APs) are among the most effective methods to model the spatial and contextual information for the analysis of remote sensing images, especially for classification task. Since their first introduction to this field in early 2010's, many research studies have been contributed not only to exploit and adapt their use to different applications, but also to extend and improve their performance for better dealing with more complex data. In this paper, we revisit and discuss different developments and extensions from APs which have drawn significant attention from researchers in the past few years. These studies are analyzed and gathered based on the concept of multi-stage AP construction. In our experiments, a comparative study on classification results of two remote sensing data is provided in order to show their significant improvements compared to the originally proposed APs.
\end{abstract}


\section{Introduction}
Image classification is one of the most crucial tasks in remote sensing imagery which serves for several applications in land use and land cover mapping and monitoring. With the emergence of high resolution remote sensing technology, the exploitation of the spatial information together with the spectral characteristics becomes more and more significant to characterize and discriminate different thematic classes present from the image content. Within such spatial-spectral context, morphological profiles (MPs) \cite{pesaresi2001new} were extensively exploited during the 2000's \cite{plaza2004new, palmason2005classification, benediktsson2005classification, fauvel2008spectral} thanks to their multilevel analysis of spatial information by applying a sequence of opening and closing by reconstruction operators with increasing-size structuring elements (SEs). However, their high computation complexity prevent them to deal with large-size images. Besides, SEs can only model the size and scale of regions without their gray-level characteristics, thus not considering contextual features such as texture and contrast. 
 
To overcome the MPs' shortcomings, morphological attribute profiles (APs) \cite{dalla2010morphological} were proposed in early 2010's as their generalization and consist in applying a sequence of attribute filters (AFs) which are more powerful than operators by reconstruction. These AFs can decompose the image according to different types of attribute (i.e. any geometric and statistical features of regions), not only restricted to the scale and size of SEs employed by MPs. Besides, the construction of APs can be efficiently implemented based on the hierarchical representation of image via tree structures (i.e. originally via min- and max-tree \cite{dalla2010morphological}), hence better dealing with large-size remote sensing images. Scalability is further ensured with parallel implementations \cite{merciol2017}.

In the past few years, a great number of research studies have been devoted to exploit and extend the use of APs applied to remote sensing image analysis, especially for classification task. These studies have been designed to improve the classification performance by focusing on the AP construction framework or adapting their use to different types of input data. In this paper, we conduct a survey on recent research studies that have been proposed and developed from the concept and application of APs. By decomposing the AP generation scheme into different stages, we regroup these studies into each specific stage in order to better provide an overview of their contribution to the general AP framework. We note that a recent survey \cite{ghamisi2015survey} also exists but its contribution has focused only on the spatial-spectral approaches using different spectral feature extraction techniques and spatial processing by the standard APs \cite{dalla2010morphological}. Our survey involves more complete and detailed investigations of different developments and extensions from APs to improve their performance and optimize their construction framework. 
\begin{figure*}
\centering 
\includegraphics[trim=20mm 65mm 20mm 60mm, clip, width=165mm]{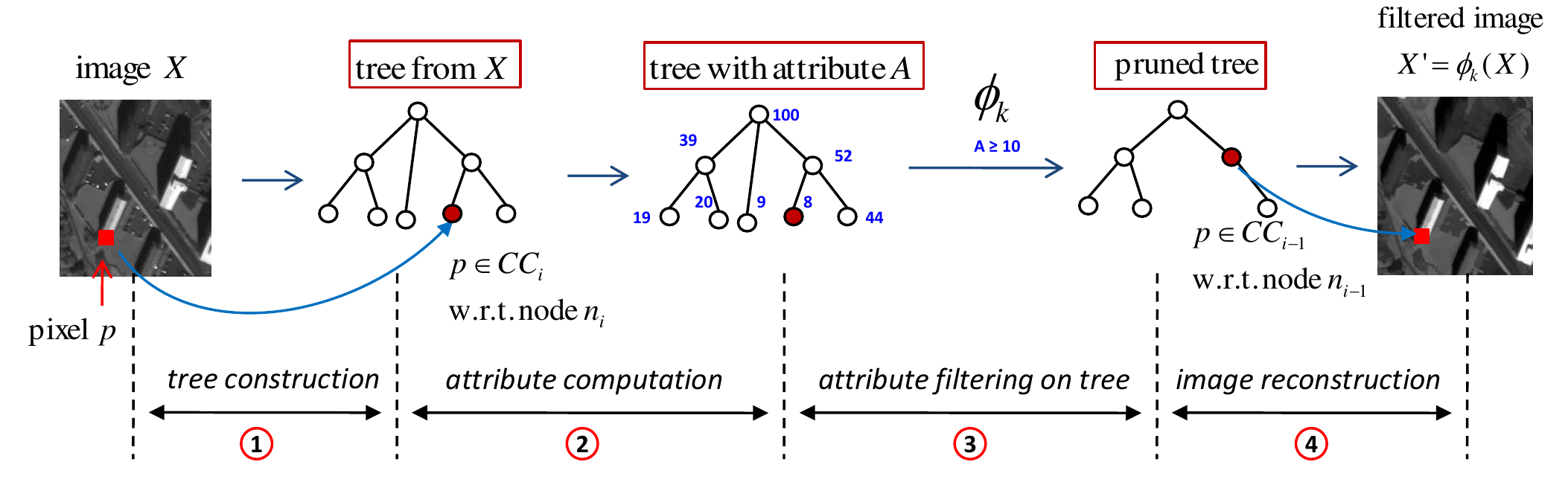}
\caption{The AP generation framework which involves four main stages: tree construction, attribute computation, tree-based attribute filtering (pruning) and image reconstruction from filtered (pruned) tree.}
\label{fig:AF}
\end{figure*}

We first recall the background of APs and highlights the key components from AP construction framework (Sec.~\ref{sec:APs}), before revisiting and discussing different developments from APs which have provided considerable contributions in the past few years (Sec.~\ref{sec:review}). An experimental study (Sec.~\ref{sec:experiments}) provides a comparative evaluation of some extensions compared to the original APs by conducting supervised classification experiments on two remote sensing image data. We finally conclude the paper and indicate future research directions (Sec.~\ref{sec:conclusion}).

\section{Principle of APs}
\label{sec:APs}
APs are multilevel image description tools obtained by successively applying a set of morphological attribute filters (AFs) \cite{dalla2010morphological}. Unlike usual image filtering operators which are directly performed on pixel level, AFs work on connected component (CC) level based on the concept of image connectivity. In other words, an AF is a filtering operator applied on CCs with regard to a specific attribute characterizing the size, shape, or other properties of objects and regions present in the image. That is why AFs are more general than operators by reconstruction (which are limited to the size and shape of SEs), and why APs are more general than MPs \cite{pesaresi2001new, dalla2010morphological}.

The generation of the standard APs \cite{dalla2010morphological} from an input image can be summarized as a four-step framework (see Fig. \ref{fig:AF}):

\begin{enumerate}
\item construct the hierarchical tree to represent the image. In \cite{dalla2010morphological}, the authors proposed to form one min-tree and one max-tree to encode the image;
\item compute some relevant attributes describing the geometrical and statistical features from each tree's node which corresponds to one connected component;
\item filter the tree by keeping/removing nodes according to their attribute values compared to predefined thresholds;
\item reconstruct the image from the filtered tree.
\end{enumerate}
Steps 3) and 4) can be done for different attributes (with different threshold values) to finally produce a set of filtered images (by stacking them) forming the final APs.

More formally, given a grayscale image $X:E \rightarrow \mathbb{Z}, E \subseteq \mathbb{Z}^2$, the standard generation of APs on X is achieved by applying a sequence of AFs based on a min-tree (attribute thickening operators $\{\phi_k\}_{k=1}^K$) and a max-tree (i.e. attribute thinning operators $\{\gamma_k\}_{k=1}^K$) as follows:
\begin{equation}
\begin{split}
\AP(X) = \Big\{\phi_K(X),\phi_{K-1}(X),\ldots,\phi_1(X),X, \\
		 \gamma_1(X),\ldots,\gamma_{K-1}(X),\gamma_K(X) \Big\},
\end{split}
\label{eq:ap}
\end{equation}
where $\phi_k(X)$ is the filtered image obtained by applying the attribute thickening $\phi$ with regard to the threshold $k$. Similar explanation is made for $\gamma_k(X)$. As observed, the resulted $\AP(X)$ is a stack of $(2K+1)$ images including the original image, $K$ filtered images from the thickening profiles and the other $K$ from the thinning profiles. 

\section{Recent advances from APs}
\label{sec:review}
As described above, the construction of APs involves four main stages which are in fact the key components that have been focused for improvements by different literature studies within the past few years. In addition, since APs basically work on panchromatic images, some pre-processing or specific adaptation procedures are required when dealing with other input data (multi-channel images, radar data, etc.). Then, spatially post-processing the output profiles to increase their description capacity for classification has also drawn attention of researchers in several research studies. 
 
We now revisit the recently proposed developments that have provided significant contributions to adapt and improve the AP framework for remote sensing image classification (Sec. \ref{subsec:input} to Sec. \ref{subsec:post}). Here, our investigation will focus on three main key features: 
\begin{itemize}
\item the adaptation of APs to other input data, in particularly to multi-channel images (Sec. \ref{subsec:input});
\item the construction of APs using various tree representation structures (Sec. \ref{subsec:tree});
\item the AP post-processing using different feature enhancement techniques (\ref{subsec:post}).
\end{itemize}
For other related extensions (Sec. \ref{subsec:attribute} and Sec. \ref{subsec:filtering}), we provide standard concepts and refer readers to the related references for further details.

\subsection{Input data}
\label{subsec:input}
Since APs were originally proposed to deal with panchromatic images \cite{dalla2010morphological}, their adaptation to other kinds of remote sensing data becomes quite significant. In particular, the application of APs to the classification of multi-channel images (multispectral and hyperspectral) has become one of the hottest research topics in this field. The idea is to perform a spatial-spectral approach for classification by combining rich spectral information from these data with efficient spatial modeling capacity of APs.

The standard extension of APs on hyperspectral images was proposed in \cite{dalla2010extended} by first applying the principal component analysis (PCA) on the image and then extracting APs from some first principal components. The advantage of PCA is that this low-complexity technique can compress most spectral information from the hyperspectral image into only some first principal components. Hence, applying APs on these components may perform a basic spectral-spatial feature extraction of the data. Other alternatives have been proposed to replace the PCA with the independent component analysis (ICA) \cite{dalla2011classification}, the kernel PCA (KPCA) \cite{bernabe2014spectral} or other supervised methods such as the discriminant analysis feature extraction (DAFE) \cite{marpu2012classification}, the non-parametric weighted feature extraction (NWFE) \cite{cavallaro2015extended}, Sparse Hilbert Schmidt Independence Criterion and surrogate kernel (HSIC) \cite{damodaran2017hsic}, etc. These methods can capture more spectral relations among hyperspectral bands and hence provide better spectral information than the PCA.

Recently, the vector strategy \cite{aptoula2016vector} has been investigated to effectively adapt APs on multispectral and hyperspectral image. The motivation of that work is to replace the \emph{marginal strategy}, i.e. independently applying APs on each image band (or each component yielded by the aforementioned feature extraction methods) and stacking them to form the extended APs, with the \emph{vector strategy} which can simultaneously process all available bands based on predefined vector-ordering relations. As a result, tree construction can be done once per multivariate image and the proposed vector APs (VAPs) become promising to deal with such hyperspectral data.

While the application of APs to optical remote sensing data has been strongly focused on, their exploitation to other remote sensing data is quite limited. One may witness some tentative work on polarimetric SAR images \cite{marpu2011spectral}, multispectral image derived features such as NDVI \cite{damodaran2017attribute} or edge information \cite{merciol2017} as well as on LiDAR data \cite{pedergnana2012classification,damodaran2017attribute}. This is still an opened topic for on-going and future research in remote sensing imagery field.

\subsection{Tree formation}
\label{subsec:tree}
Tree formation is the first principal stage of the AP construction framework (Fig. \ref{fig:AF}). As described in Sec. \ref{sec:APs}, the standard APs \cite{dalla2010morphological} were computed based on one max-tree and one min-tree (i.e. both are component trees). Other work has been proposed to exploit the inclusion tree (i.e. tree of shapes) \cite{dalla2011self} in order to form the self-dual APs (SDAPs). The advantages of using such a tree of shapes are twofold. First, its self-dual property enables the attribute filtering operators to simultaneously access and model both dark and bright regions from the image. And secondly, by using only one tree of shapes to replace both min-tree and max-tree \cite{dalla2010morphological}, the feature dimension of SDAPs is reduced to half of that of APs. Consequently, SDAPs have been proved to be more efficient than APs in many research studies \cite{cavallaro2015extended, dalla2011self, cavallaro2016remote}. 

Since the above component and inclusion trees both rely on an ordering relation of the image pixels, their construction from multivariate images (e.g. multi- and hyperspectral data) is not straightforward. That is why the authors in \cite{bosilj2017attribute} have recently investigated and proposed to use the partition trees such as $\alpha$-tree and $\omega$-tree to compute the $\alpha$-APs, $\omega$-APs, respectively. These profiles have been proved to provide fair performance compared to the standard APs. Moreover, they offer the possibility to work on multivariate images only using a single tree. Furthermore, it is also possible to rely on training samples to perform metric learning so as to provide the basic elements required for a partitioning tree \cite{lefevre2014hyperspectral}.

\subsection{Node attributes and threshold selection}
\label{subsec:attribute}
The selection of tree node attributes as well as their thresholds for filtering on tree plays also an important role. Node attributes are usually related to the geometrical (such as size, shape) and statistical features (pixel distribution, texture, etc.) of the CC corresponding to the node. In the literature, four attributes have been used in most studies related to remote sensing image classification: \emph{area}, \emph{standard deviation}, \emph{moment of inertia}, \emph{diagonal length of bounding box}.

After deciding which attributes to calculate from nodes, the setting of their threshold values has been also concerned. Early work \cite{dalla2010morphological, dalla2011classification, dalla2011self} usually set attribute thresholds manually based on experiments on some specific image data. However, since those values might be not applicable to other data, automatic threshold selection has drawn attention from many researchers. Some interesting studies have been proposed to automatically compute attribute thresholds using fixed formulas \cite{marpu2013automatic, ghamisi2014automatic}, supervised approaches \cite{mahmood2012automatic, pedergnana2013novel} as well as granulometric characteristic functions \cite{cavallaro2015automatic, cavallaro2017automatic}. Readers are referred to the mentioned papers for further details about these attribute selection strategies.

\subsection{Tree filtering}
\label{subsec:filtering}
Once the tree is formed and the attributes together with their thresholds are selected, the next stage is to evaluate each node in order to filter (i.e prune) the tree. Basically, there are two filtering rules including the \emph{pruning strategy} (min, max, Viterbi decision rules) and the \emph{non-pruning strategy} (direct, subtractive rules) \cite{dalla2010morphological}. Studies on the effect of different filtering rules have been done by \cite{cavallaro2014comparison, cavallaro2016remote}.

\subsection{Post-processing of output profiles}
\label{subsec:post}
The output AP features, i.e. sequence of filtered images in Eq. \eqref{eq:ap}, can be directly fed into supervised classifiers such as SVM or Random Forest for classification task. Such direct application has provided better performance compared to MPs \cite{pesaresi2001new} in terms of classification accuracy as well as computational cost. However, since APs still involve quite redundant information within their high-dimension features, the post-processing of these profiles to improve their performance has been addressed in several studies. First and foremost, many studies have proposed to apply different feature selection techniques on APs to extract highly informative features and reduce their dimension. In \cite{marpu2012classification, ghamisi2014automatic, bernabe2014spectral}, both linear (PCA, ICA) and nonlinear methods (ICA, KPCA, DAFE, DBFE, NWFE, etc.) have been investigated. A general framework as well as a systematic survey on spatial-spectral approaches combining APs with these feature selection techniques have been investigated in \cite{ghamisi2015survey}.

Other work has focused on extra spatial processing of APs for better characterization of structural and textural information from the image content. Recent studies believe that when dealing with VHR remote sensing images from which regions and objects become more heterogeneous, APs may not provide a complete spatial characterization of pixels. Therefore, some efforts have been proposed to replace each AP sample response by the histogram or some first-order statistical features of the local patch around that AP's pixel position. As a result, the local histogram-based APs (HAPs) \cite{demir2016histogram, battiti2015compressed} and the local feature-based APs (LFAPs) \cite{pham2017local,pham2017classification} have been proposed and proved to be more efficient for better dealing with local textures. Then, the extensions of these extra spatial processing methods on the self-dual profiles (using the tree of shapes) as well as on hyperspectral images have been provided \cite{pham2017local}.

Last but not least, we refer readers to some other frameworks using the sparse representation \cite{song2014remotely} or the deep learning approach \cite{aptoula2016deep} for post-processing of AP features. Also, some ensemble methods \cite{xia2015random, bao2016combining} have been applied to better exploit and combine AP features to improve the classification performance.

\section{Experimental study}
\label{sec:experiments}
This section describes our experimental study to evaluate the performance of the standard APs as well as some of their improvements and extensions. Supervised classification has been carried out on both panchromatic and hyperspectral image data in order to provide a comparative study. We first introduce the two data sets and the experimental setup. Then, classification results will be provided.

\subsection{Data description}
\subsubsection{Reykjavik data set}
The first data set is a panchromatic image of size $628 \times 700$ pixels acquired by the IKONOS Earth imaging satellite with 1-m resolution in Reykjavik, Iceland. This data consists of six thematic classes including residential, soil, shadow, commercial, highway and road. The image was provided with already-split training and test sets (22741 training samples and 98726 test samples). The input image together with its thematic ground truth map for testing and training sets are shown in Fig. \ref{fig:dataset}(a).

\subsubsection{Pavia University data set} 
\begin{figure}[!ht]
\centerline{\begin{minipage}[b]{0.94\linewidth}
  \centering
  \includegraphics[width=26mm]{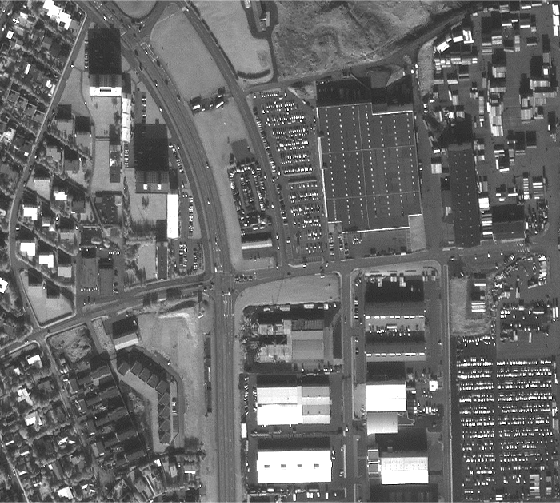}
  \hfill
  \includegraphics[width=26mm]{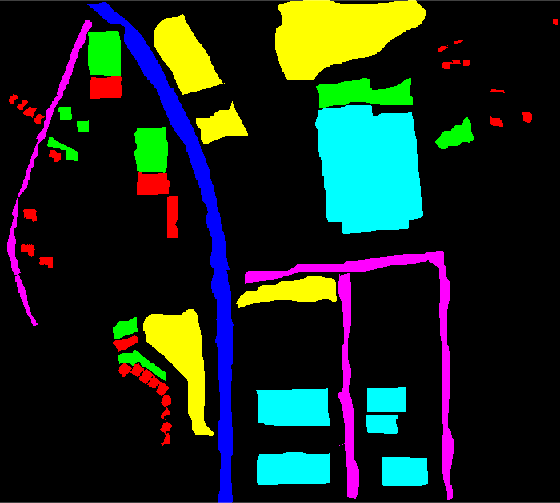}
  \hfill
  \includegraphics[width=26mm]{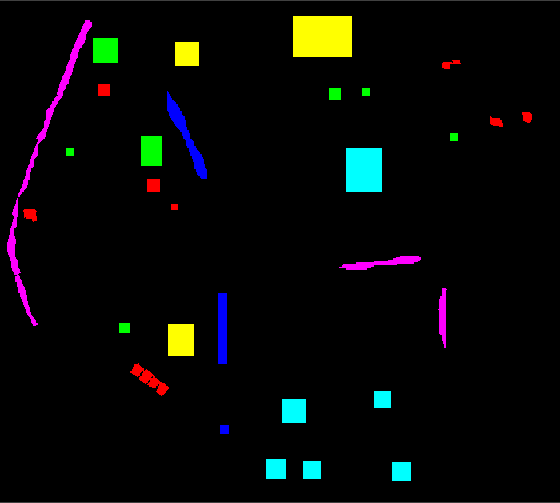} 
  \vfill
  \vspace{1mm}
  \includegraphics[trim=35mm 87mm 30mm 72mm, clip, width=85mm]{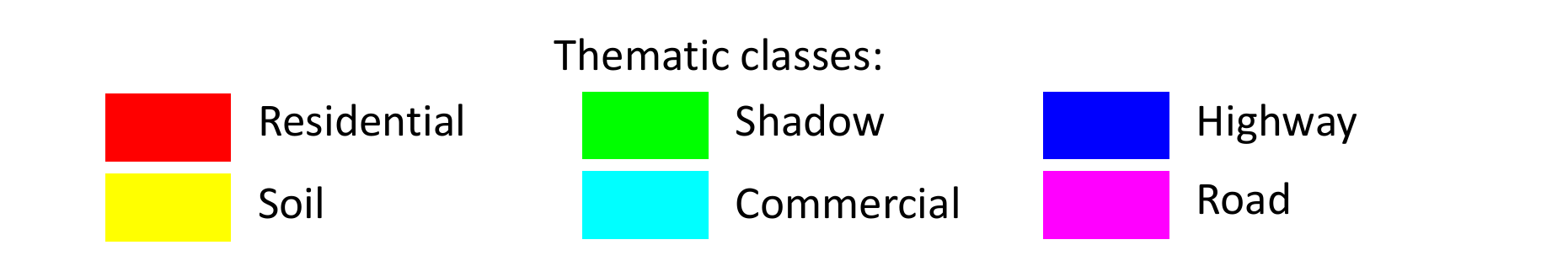}
  \vfill
  \footnotesize{\textbf{(a)}}             
\end{minipage}}
\vfill
\vspace{2mm}
\centerline{\begin{minipage}[b]{0.94\linewidth}
  \centering
  \includegraphics[width=26mm]{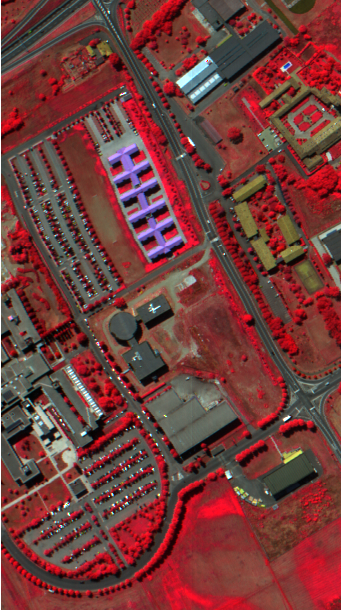}
  \hfill
  \includegraphics[width=26mm]{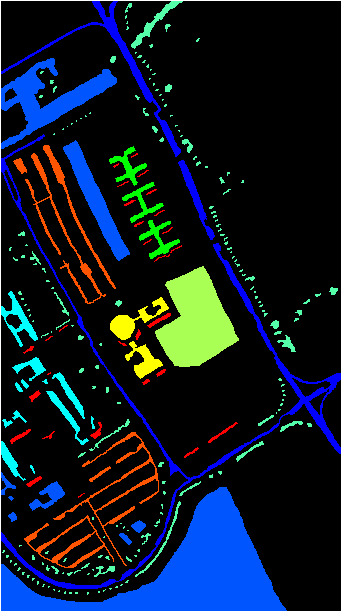}
  \hfill
  \includegraphics[width=26mm]{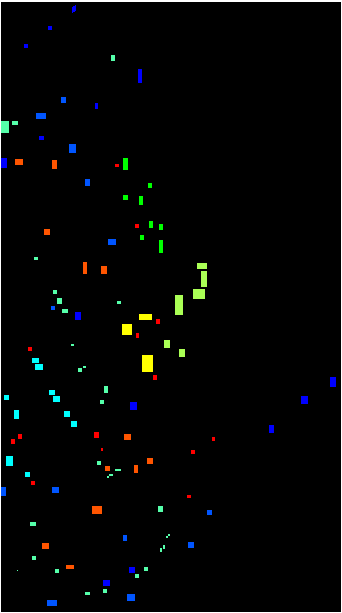} 
  \vfill
  \vspace{1mm}
  \includegraphics[trim=40mm 76mm 30mm 72mm, clip, width=85mm]{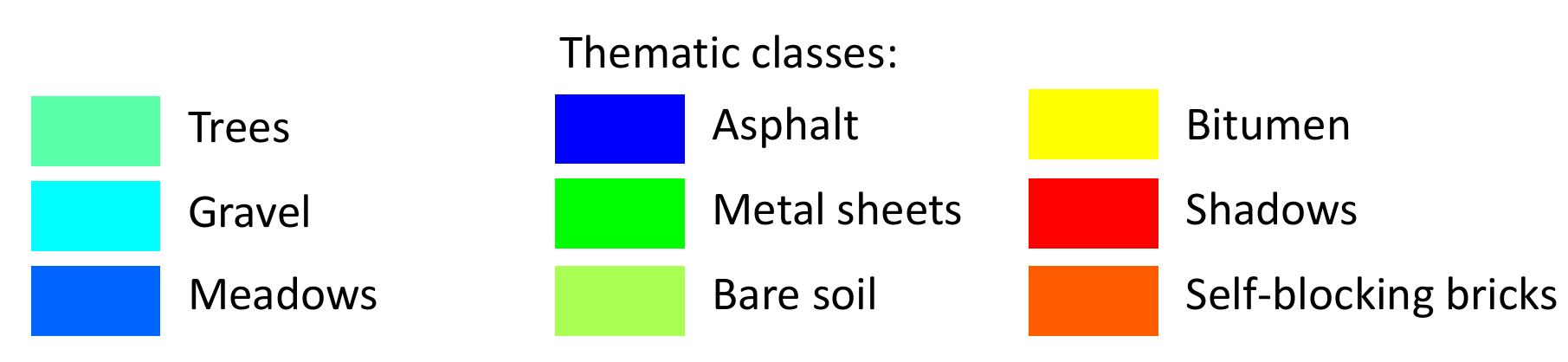}
  \vfill
  \footnotesize{\textbf{(b)}}             
\end{minipage}}
\caption{Two data sets used in our experimental study. \textbf{(a)} The $628\times 700$ Reykjavik data (left to right: panchromatic, thematic ground truth with 6 classes and training set); \textbf{(b)} The $610\times 340$ Pavia University data (left to right: false-color image made by bands 31-56-102, ground truth including nine thematic classes and training set).}
\label{fig:dataset}
\end{figure}

The second data set is the hyperspectral image acquired by the ROSIS airborne sensor with 1.3-m spatial resolution over the region of Pavia University, Italy. The image consists of $610 \times 340$ pixels with 103 spectral bands (from 0.43 to 0.86 $\mu$m) and covers nine thematic classes: trees, asphalt, bitumen, gravel, metal sheets, shadows, meadows, self-blocking bricks and bare soil. For this image, 3921 training samples and 42776 test samples were split for classification task. The false-color image (made by combining the bands 31, 56 and 102), the ground truth map and the training set are shown in Fig. \ref{fig:dataset}(b). As previously discussed, for this data set, we first performed the PCA on the image and the first four PCs (involving more than $99\%$ of the total variance) were preserved for our experiments.

\subsection{Setup}
Supervised classification was conducted on the two data sets using the random forest classifier \cite{liaw2002classification} with 100 trees. The number of variables involved in the training was set to the square root of the feature vector length. In order to evaluate and compare classification accuracy of different approaches, overall accuracy (OA), average accuracy (AA), and kappa coefficient ($\kappa$) have been taken into account. For attribute filtering, we exploited two attributes including the \emph{area} and the \emph{moment of inertia}. Ten area thresholds were adopted for the Reykjavik data as proposed by several papers \cite{cavallaro2016remote,pham2017feature,ghamisi2016extinction}. For the Pavia University data, fourteen thresholds were automatically computed according to \cite{ghamisi2014automatic}. We have:
\begin{equation*}
\begin{split}
\lambda_{a,Rey} =
\{25, 100, 500, 1000, 5000, 10 000,  \\
	20 000, 50 000, 100 000, 150 000\},
\end{split}
\end{equation*}
\begin{equation*}
\begin{split}
\lambda_{a,Pav} =
\{770, 1538, 2307, 3076, 3846, 4615, 5384, \\
		 6153, 6923, 7692, 8461, 9230, 10000, 10769\}.
\end{split}
\end{equation*}
Next, the manual settings used in many studies \cite{dalla2010extended, cavallaro2015extended, aptoula2016vector} were adopted for the moment of inertia attribute as follows:  
$$\lambda_{i,Rey} = \lambda_{i,Pav} = \{0.2, 0.3, 0.4, 0.5\}.$$

In the following subsection, we report and compare the classification results yielded by the APs generated from different kinds of tree including: the max-tree (AP-maxT), the min-tree (AP-minT), one max-tree and one min-tree (standard APs) \cite{dalla2010morphological}, the SDAPs \cite{dalla2011self}, the $\alpha$-APs and $\omega$-APs \cite{bosilj2017attribute}. We also provide the results of some effective post-processing techniques including the HAPs/HSDAPs \cite{demir2016histogram}, LFAPs/LFSDAPs \cite{pham2017local} and the deep learning approach (deep-APs) \cite{aptoula2016deep}. Then, for the hyperspectral Pavia data, VAPs \cite{aptoula2016vector} are evaluated as well. Here, we perform standard implementation as well as equivalent parameter configuration of these methods to ensure a fair comparison.

\subsection{Results}
Tables \ref{tab:result_reykjavik} and \ref{tab:result_pavia} report the classification results of the Reykjavik and the Pavia data, respectively, yielded by the above mentioned methods. The calculation of each method's feature dimension can be consulted from the related papers. Here, we provide some remarks in terms of classification performance. For both data sets, we observe that those extension methods can provide extra classification accuracy compared to the standard APs but behave differently for each image. 
\begin{table}[!ht]
	\centering
	\caption{Classification result of the Reykjavik data obtained by different methods using Random Forest with 100 trees.}
	\label{tab:result_reykjavik}
	\resizebox{0.45\textwidth}{!}{%
	{\renewcommand{\arraystretch}{1.05}
	\begin{tabu}{l| c| c c c}
	\hline
	\hline
	\multirow{2}{*}{\textbf{Method}} & \multirow{2}{*}{\textbf{Dimension}} & \multicolumn{3}{c}{\textbf{Classification result}} \\
	\cline{3-5}
	 & & OA ($\%$) & AA ($\%$) & $\kappa$ \\
	 \hline
	PAN & 1 & 63.21 & 53.58 & 0.5237 \\
	AP-maxT & 16 & 73.31 & 68.23 & 0.6597\\
	AP-minT & 16 & 72.37 & 64.63 & 0.6449\\
	AP & 30 & 82.02 & 78.42 & 0.7730 \\
	\hline
	$\alpha$-AP & 16 & 77.38 & 70.19 & 0.7101 \\
	$\omega$-AP & 16 & 76.68 & 70.23 & 0.7024 \\
	SDAP & 16 & 86.06 & 82.36 & 0.8237 \\
	\hline
	HAP & 180 & 84.67 & 81.89 & 0.8055 \\
	HSDAP & 96 & 86.05 & 81.67 & 0.8234 \\
	LFAP & 60 & 87.44 & 85.21 & 0.8411 \\
	LFSDAP & 32 & 89.17 & 87.08 & 0.8631 \\
	Deep-AP & 1024 & 86.09 & 83.01 & 0.8230 \\
	\hline
	\end{tabu} 
	}
	}
\end{table}

For Reykjavik image, efforts on changing the tree formation have provided some considerable effects. Indeed, the $\alpha$-APs and $\omega$-APs could outperform APs on each single max-tree or min-tree but still falls below the standard APs. Then, by using the tree of shapes, SDAPs significantly improved the accuracy with approximately $4\%$ in OA ($86.06\%$ compared to $82.02\%$) and $5\%$ in $\kappa$ ($0.824$ compared to $0.773$). Next, by post-processing the output profiles, techniques like HAP, LFAP and deep-AP have also provided important improvements. Consequently, the best classification result was obtained by using the local feature-based profiles  with $\OA=87.44\%$ ($\kappa=0.841$) using min-tree and max-tree (LFAP) and $\OA=89.17\%$ ($\kappa=0.863$) using the tree of shapes (LFSDAP). Compared to the standard APs, an OA enhancement of $5.42\%$ and $7.15\%$, respectively, was achieved. 
\begin{table}[!ht]
	\centering
	\caption{Classification result of the Pavia University data obtained by different methods using Random Forest with 100 trees.}
	\label{tab:result_pavia}
	\resizebox{0.45\textwidth}{!}{%
	{\renewcommand{\arraystretch}{1.05}
	\begin{tabu}{l| c| c c c}
	\hline
	\hline
	\multirow{2}{*}{\textbf{Method}} & \multirow{2}{*}{\textbf{Dimension}} & \multicolumn{3}{c}{\textbf{Classification result}} \\
	\cline{3-5}
	 & & OA ($\%$) & AA ($\%$) & $\kappa$ \\
	 \hline
	4 PCs & 4 & 70.62 & 80.37 & 0.6375 \\
	AP-maxT & 80 & 83.66 & 88.52 & 0.7936\\
	AP-minT & 80 & 81.48 & 86.11 & 0.7596\\
	AP & 152 & 91.66 & 93.96 & 0.8891 \\
	\hline
	SDAP & 80 & 94.28 & 93.96 & 0.9234\\
	$\alpha$-AP & 80 & 94.52 & 94.11 & 0.9293 \\
	$\omega$-AP & 80 & 96.10 & 95.66 & 0.9403 \\
	\hline
	HAP & 912 & 94.14 & 94.40 & 0.9234\\
	HSDAP & 480 & 94.53 & 92.64 & 0.9266\\
	LFAP & 304 & 93.57 & 93.50 & 0.9149\\
	LFSDAP & 160 & 95.25 & 94.49 & 0.9363\\
	VAP & 152 & 96.30 & 95.64 & 0.9500\\
	Deep-AP & 1024 & 99.02 & 98.54 & 0.9790 \\
	\hline
	\end{tabu} 
	}
	}
\end{table}

For the hyperspectral Pavia data, we observe that APs built from different tree structures yielded different behaviors compared to the Reykjavik image. This time, the $\alpha$-APs and $\omega$-APs outperformed both APs and SDAPs. In particular, by using the $\omega$-tree, one can achieve an $\OA=96.10\%$, i.e. $4.44\%$ and $1.82\%$ better than standard APs and SDAPs, respectively. For post-processing methods, VAPs and deep-APs provided better performance compared to HAPs and LFAPs. These methods have been proved to be efficient within a spatial-spectral context usually applied to hyperspectral data. As a result, the best classification accuracy was achieved by deep-APs with $\OA=99.02\%$ and $\kappa=0.979$. Compared to the standard APs, an enhancement of $7.36\%$ in OA and $9\%$ in $\kappa$ was adopted.

\section{Conclusion}
\label{sec:conclusion}
We have conducted a survey on recent developments from morphological attribute profiles in the context of remote sensing image classification. Three key components have been focused including the AP adaptation on multi-channel image data, the use of different tree representations and the various AP post-processing procedures. Experimental study on one panchromatic and one hyperspectral image has been performed to provide a general evaluation of different methods compared to the original framework. This paper may serve as an overview of AP recent advances to readers as well as a guidance to researchers working on this framework and its alternatives within their work. We believe the exploitation and adaptation of APs in remote sensing imagery still remains an open research topic for on-going as well as future work.

\section*{Acknowledgment}
This work was supported the R\'egion Bretagne grant and the BAGEP Award of the Science Academy and the Tubitak grant 115E857. 
The authors would like to thank Prof. Jon Atli Benediktsson and Prof. Paolo Gamba for making available the Reykjavik image and the hyperspectral Pavia University data.

%% ============ REFERENCES =============
\bibliographystyle{IEEEtran}
\bibliography{RefAbrv,RefAPs}

\begin{thebibliography}{10}
\providecommand{\url}[1]{#1}
\csname url@samestyle\endcsname
\providecommand{\newblock}{\relax}
\providecommand{\bibinfo}[2]{#2}
\providecommand{\BIBentrySTDinterwordspacing}{\spaceskip=0pt\relax}
\providecommand{\BIBentryALTinterwordstretchfactor}{4}
\providecommand{\BIBentryALTinterwordspacing}{\spaceskip=\fontdimen2\font plus
\BIBentryALTinterwordstretchfactor\fontdimen3\font minus
  \fontdimen4\font\relax}
\providecommand{\BIBforeignlanguage}[2]{{%
\expandafter\ifx\csname l@#1\endcsname\relax
\typeout{** WARNING: IEEEtran.bst: No hyphenation pattern has been}%
\typeout{** loaded for the language `#1'. Using the pattern for}%
\typeout{** the default language instead.}%
\else
\language=\csname l@#1\endcsname
\fi
#2}}
\providecommand{\BIBdecl}{\relax}
\BIBdecl

\bibitem{pesaresi2001new}
M.~Pesaresi and J.~A. Benediktsson, ``A new approach for the morphological
  segmentation of high-resolution satellite imagery,'' \emph{{IEEE} Trans.
  Geosci. Remote Sens.}, vol.~39, no.~2, pp. 309--320, 2001.

\bibitem{plaza2004new}
A.~Plaza, P.~Martinez, R.~Perez, and J.~Plaza, ``A new approach to mixed pixel
  classification of hyperspectral imagery based on extended morphological
  profiles,'' \emph{Pattern Recogn.}, vol.~37, no.~6, pp. 1097--1116, 2004.

\bibitem{palmason2005classification}
J.~A. Palmason, J.~A. Benediktsson, J.~R. Sveinsson, and J.~Chanussot,
  ``Classification of hyperspectral data from urban areas using morphological
  preprocessing and independent component analysis,'' in \emph{Proc. {IEEE}
  IGARSS}, vol.~1, 2005, pp. 176--179.

\bibitem{benediktsson2005classification}
J.~A. Benediktsson, J.~A. Palmason, and J.~R. Sveinsson, ``Classification of
  hyperspectral data from urban areas based on extended morphological
  profiles,'' \emph{{IEEE} Trans. Geosci. Remote Sens.}, vol.~43, no.~3, pp.
  480--491, 2005.

\bibitem{fauvel2008spectral}
M.~Fauvel, J.~A. Benediktsson, J.~Chanussot, and J.~R. Sveinsson, ``Spectral
  and spatial classification of hyperspectral data using svms and morphological
  profiles,'' \emph{{IEEE} Trans. Geosci. Remote Sens.}, vol.~46, no.~11, pp.
  3804--3814, 2008.

\bibitem{dalla2010morphological}
M.~Dalla~Mura, J.~A. Benediktsson, B.~Waske, and L.~Bruzzone, ``Morphological
  attribute profiles for the analysis of very high resolution images,''
  \emph{{IEEE} Trans. Geosci. Remote Sens.}, vol.~48, no.~10, pp. 3747--3762,
  2010.

\bibitem{merciol2017}
F.~Merciol, T.~Balem, and S.~Lef{\`e}vre, ``Efficient and large-scale land
  cover classification using multiscale image analysis,'' in \emph{ESA Conf. on
  Big Data from Space (BiDS)}, 2017.

\bibitem{ghamisi2015survey}
P.~Ghamisi, M.~Dalla~Mura, and J.~A. Benediktsson, ``A survey on
  spectral--spatial classification techniques based on attribute profiles,''
  \emph{{IEEE} Trans. Geosci. Remote Sens.}, vol.~53, no.~5, pp. 2335--2353,
  2015.

\bibitem{dalla2010extended}
M.~Dalla~Mura, J.~Atli~Benediktsson, B.~Waske, and L.~Bruzzone, ``Extended
  profiles with morphological attribute filters for the analysis of
  hyperspectral data,'' \emph{Int. J. Remote Sens.}, vol.~31, no.~22, pp.
  5975--5991, 2010.

\bibitem{dalla2011classification}
M.~Dalla~Mura, A.~Villa, J.~A. Benediktsson, J.~Chanussot, and L.~Bruzzone,
  ``Classification of hyperspectral images by using extended morphological
  attribute profiles and independent component analysis,'' \emph{{IEEE} Geosci.
  Remote Sens. Lett.}, vol.~8, no.~3, pp. 542--546, 2011.

\bibitem{bernabe2014spectral}
S.~Bernabe, P.~R. Marpu, A.~Plaza, M.~Dalla~Mura, and J.~A. Benediktsson,
  ``Spectral--spatial classification of multispectral images using kernel
  feature space representation,'' \emph{{IEEE} Geosci. Remote Sens. Lett.},
  vol.~11, no.~1, pp. 288--292, 2014.

\bibitem{marpu2012classification}
P.~R. Marpu, M.~Pedergnana, M.~Dalla~Mura, S.~Peeters, J.~A. Benediktsson, and
  L.~Bruzzone, ``Classification of hyperspectral data using extended attribute
  profiles based on supervised and unsupervised feature extraction
  techniques,'' \emph{Int. J. Image Data Fusion}, vol.~3, no.~3, pp. 269--298,
  2012.

\bibitem{cavallaro2015extended}
G.~Cavallaro, M.~Dalla~Mura, J.~A. Benediktsson, and L.~Bruzzone, ``Extended
  self-dual attribute profiles for the classification of hyperspectral
  images,'' \emph{{IEEE} Geosci. Remote Sens. Lett.}, vol.~12, no.~8, pp.
  1690--1694, 2015.

\bibitem{damodaran2017hsic}
B.~B. Damodaran, N.~Courty, and S.~Lef\`evre, ``Sparse hilbert schmidt
  independence criterion and surrogate-kernel-based feature selection for
  hyperspectral image classification,'' \emph{{IEEE} Trans. Geosci. Remote
  Sens.}, vol.~55, no.~4, pp. 2385--2398, 2017.

\bibitem{aptoula2016vector}
E.~Aptoula, M.~Dalla~Mura, and S.~Lef{\`e}vre, ``Vector attribute profiles for
  hyperspectral image classification,'' \emph{{IEEE} Trans. Geosci. Remote
  Sens.}, vol.~54, no.~6, pp. 3208--3220, 2016.

\bibitem{marpu2011spectral}
P.~R. Marpu, K.-S. Chen, C.-Y. Chu, and J.~A. Benediktsson, ``Spectral-spatial
  classification of polarimetric {SAR} data using morphological profiles,'' in
  \emph{Synthetic Aperture Radar (APSAR), 2011 3rd Int. Asia-Pacific Conf.},
  2011, pp. 1--3.

\bibitem{damodaran2017attribute}
B.~B. Damodaran, J.~H{\"o}hle, and S.~Lef{\`e}vre, ``Attribute profiles on
  derived features for urban land cover classification,'' \emph{Photogrammetric
  Engineering \& Remote Sensing}, vol.~83, no.~3, pp. 183--193, 2017.

\bibitem{pedergnana2012classification}
M.~Pedergnana, P.~R. Marpu, M.~Dalla~Mura, J.~A. Benediktsson, and L.~Bruzzone,
  ``Classification of remote sensing optical and lidar data using extended
  attribute profiles,'' \emph{IEEE J. Sel. Topics Sig. Proc.}, vol.~6, no.~7,
  pp. 856--865, 2012.

\bibitem{dalla2011self}
M.~Dalla~Mura, J.~Benediktsson, and L.~Bruzzone, ``Self-dual attribute profiles
  for the analysis of remote sensing images,'' in \emph{Int. Symp. Math.
  Morpho. Appl. Sig. Image Proc.}, 2011, pp. 320--330.

\bibitem{cavallaro2016remote}
G.~Cavallaro, M.~Dalla~Mura, J.~A. Benediktsson, and A.~Plaza, ``Remote sensing
  image classification using attribute filters defined over the tree of
  shapes,'' \emph{{IEEE} Trans. Geosci. Remote Sens.}, vol.~54, no.~7, pp.
  3899--3911, 2016.

\bibitem{bosilj2017attribute}
P.~Bosilj, B.~B. Damodaran, E.~Aptoula, M.~Dalla~Mura, and S.~Lef{\`e}vre,
  ``Attribute profiles from partitioning trees,'' in \emph{Int. Symp. Math.
  Morpho. Its Appl. Sig. Image Proc.}, 2017, pp. 381--392.

\bibitem{lefevre2014hyperspectral}
S.~Lef\`evre, L.~Chapel, and F.~Merciol, ``Hyperspectral image classification
  from multiscale description with constrained connectivity and metric
  learning,'' in \emph{IEEE Wksh. Hyper. Image and Sig. Proc.: Evol. Remote
  Sens. (WHISPERS)}, 2014, pp. 1--4.

\bibitem{marpu2013automatic}
P.~R. Marpu, M.~Pedergnana, M.~Dalla~Mura, J.~A. Benediktsson, and L.~Bruzzone,
  ``Automatic generation of standard deviation attribute profiles for
  spectral--spatial classification of remote sensing data,'' \emph{{IEEE}
  Geosci. Remote Sens. Lett.}, vol.~10, no.~2, pp. 293--297, 2013.

\bibitem{ghamisi2014automatic}
P.~Ghamisi, J.~A. Benediktsson, and J.~R. Sveinsson, ``Automatic
  spectral--spatial classification framework based on attribute profiles and
  supervised feature extraction,'' \emph{{IEEE} Trans. Geosci. Remote Sens.},
  vol.~52, no.~9, pp. 5771--5782, 2014.

\bibitem{mahmood2012automatic}
Z.~Mahmood, G.~Thoonen, and P.~Scheunders, ``Automatic threshold selection for
  morphological attribute profiles,'' in \emph{Proc. {IEEE} IGARSS}, 2012, pp.
  4946--4949.

\bibitem{pedergnana2013novel}
M.~Pedergnana, P.~R. Marpu, M.~Dalla~Mura, J.~A. Benediktsson, and L.~Bruzzone,
  ``A novel technique for optimal feature selection in attribute profiles based
  on genetic algorithms,'' \emph{{IEEE} Trans. Geosci. Remote Sens.}, vol.~51,
  no.~6, pp. 3514--3528, 2013.

\bibitem{cavallaro2015automatic}
G.~Cavallaro, N.~Falco, M.~Dalla~Mura, L.~Bruzzone, and J.~A. Benediktsson,
  ``Automatic threshold selection for profiles of attribute filters based on
  granulometric characteristic functions,'' in \emph{Int. Symp. Math. Morpho.
  Its Appl. Sig. Image Proc.}, 2015, pp. 169--181.

\bibitem{cavallaro2017automatic}
G.~Cavallaro, N.~Falco, M.~Dalla~Mura, and J.~A. Benediktsson, ``Automatic
  attribute profiles,'' \emph{{IEEE} Trans. Image Processing}, vol.~26, no.~4,
  pp. 1859--1872, 2017.

\bibitem{cavallaro2014comparison}
G.~Cavallaro, M.~Dalla~Mura, J.~A. Benediktsson, and L.~Bruzzone, ``A
  comparison of self-dual attribute profiles based on different filter rules
  for classification,'' in \emph{Proc. {IEEE} IGARSS}, 2014, pp. 1265--1268.

\bibitem{demir2016histogram}
B.~Demir and L.~Bruzzone, ``Histogram-based attribute profiles for
  classification of very high resolution remote sensing images,'' \emph{{IEEE}
  Trans. Geosci. Remote Sens.}, vol.~54, no.~4, pp. 2096--2107, 2016.

\bibitem{battiti2015compressed}
R.~Battiti, B.~Demir, and L.~Bruzzone, ``Compressed histogram attribute
  profiles for the classification of {VHR} remote sensing images,'' in
  \emph{SPIE Remote Sens.}, 2015, pp. 96\,430R--96\,430R.

\bibitem{pham2017local}
M.-T. Pham, S.~Lef{\`e}vre, and E.~Aptoula, ``Local feature-based attribute
  profiles for optical remote sensing image classification,'' \emph{{IEEE}
  Trans. Geosci. Remote Sens.}, vol.~56, no.~2, pp. 1199--1212, 2018.

\bibitem{pham2017classification}
M.-T. Pham, S.~Lef{\`e}vre, E.~Aptoula, and B.~B. Damodaran, ``Classification
  of {VHR} remote sensing images using local feature-based attribute
  profiles,'' in \emph{Proc. {IEEE} IGARSS}, 2017, pp. 1083--1086.

\bibitem{song2014remotely}
B.~Song, J.~Li, M.~Dalla~Mura, P.~Li, A.~Plaza, J.~M. Bioucas-Dias, J.~A.
  Benediktsson, and J.~Chanussot, ``Remotely sensed image classification using
  sparse representations of morphological attribute profiles,'' \emph{{IEEE}
  Trans. Geosci. Remote Sens.}, vol.~52, no.~8, pp. 5122--5136, 2014.

\bibitem{aptoula2016deep}
E.~Aptoula, M.~C. Ozdemir, and B.~Yanikoglu, ``Deep learning with attribute
  profiles for hyperspectral image classification,'' \emph{{IEEE} Geosci.
  Remote Sens. Lett.}, vol.~13, no.~12, pp. 1970--1974, 2016.

\bibitem{xia2015random}
J.~Xia, M.~Dalla~Mura, J.~Chanussot, P.~Du, and X.~He, ``Random subspace
  ensembles for hyperspectral image classification with extended morphological
  attribute profiles,'' \emph{{IEEE} Trans. Geosci. Remote Sens.}, vol.~53,
  no.~9, pp. 4768--4786, 2015.

\bibitem{bao2016combining}
R.~Bao, J.~Xia, M.~Dalla~Mura, P.~Du, J.~Chanussot, and J.~Ren, ``Combining
  morphological attribute profiles via an ensemble method for hyperspectral
  image classification,'' \emph{{IEEE} Geosci. Remote Sens. Lett.}, vol.~13,
  no.~3, pp. 359--363, 2016.

\bibitem{liaw2002classification}
A.~Liaw, M.~Wiener \emph{et~al.}, ``Classification and regression by
  {randomForest},'' \emph{R news}, vol.~2, no.~3, pp. 18--22, 2002.

\bibitem{pham2017feature}
M.-T. Pham, E.~Aptoula, and S.~Lef{\`e}vre, ``Feature profiles from attribute
  filtering for remote sensing image classification,'' \emph{{IEEE} J. Sel.
  Topics Appl. Earth Observations Remote Sens.}, vol.~11, no.~1, pp. 249--256,
  2018.

\bibitem{ghamisi2016extinction}
P.~Ghamisi, R.~Souza, J.~A. Benediktsson, X.~X. Zhu, L.~Rittner, and R.~A.
  Lotufo, ``Extinction profiles for the classification of remote sensing
  data,'' \emph{{IEEE} Trans. Geosci. Remote Sens.}, vol.~54, no.~10, pp.
  5631--5645, 2016.

\end{thebibliography}

\end{document}